\documentclass[11pt]{article}

\usepackage{acl}
\usepackage{times}
\usepackage{latexsym}
\usepackage[T1]{fontenc}
\usepackage[utf8]{inputenc}
\usepackage{microtype}
\usepackage{fontspec}

\usepackage{inconsolata}
\usepackage{graphicx}

\usepackage{listings}
\usepackage[T1]{fontenc}        
\usepackage{textcomp}           
\title{Retriv at BLP-2025 Task 2: Test-Driven Feedback-Guided Framework for Bangla-to-Python Code Generation }

\author{
    K M Nafi Asib, Sourav Saha, and Mohammed Moshiul Hoque \\
    Department of Computer Science and Engineering \\
    Chittagong University of Engineering and Technology \\
    \{nafi.asib, sahasourav1170\}@gmail.com \\
    moshiul\_240@cuet.ac.bd
}

\begin{document}
\maketitle
\begin{abstract}

Large Language Models (LLMs) have advanced the automated generation of code from natural language prompts. However, low-resource languages (LRLs) like Bangla remain underrepresented due to the limited availability of instruction-to-code datasets and evaluation benchmarks. To address this, the BLP Workshop at IJCNLP-AACL 2025 introduced a shared task on ``Code Generation in Bangla''. In this work, we propose a method that combines instruction prompting with a test-driven, feedback-guided iterative refinement process using a fine-tuned Qwen2.5-14B model. The model generates code from Bangla instructions, tests it against unit tests, and iteratively refines any failing outputs through three evaluation passes, using test feedback to guide each step. This approach helped our team ``Retriv'' to secure 2\textsuperscript{nd} place in the shared task with a Pass@1 score of 0.934. The analysis highlights challenges in Bangla instruction understanding and Python code generation, emphasizing the need for targeted methods in LRLs. We made experimental scripts publicly available for the community.\footnote{\url{https://github.com/NafiAsib/Retriv-BLP25-Task-2}}

\end{abstract}

\section{Introduction}
Automated code generation from natural language has made significant progress with Large Language Models (LLMs), which generate code snippets tailored to meet user needs. These models, trained on millions of open-source code repositories, perform best in high-resource languages such as English, where large, aligned datasets and benchmarks are available. In contrast, low-resource languages such as Bangla have received less attention due to the lack of high-quality instruction-to-code datasets~\cite{raihan-etal-2025-mhumaneval}. Recent work, including the TigerCoder models~\cite{raihan2025tigercoder} and benchmarks such as mHumanEval~\cite{raihan-etal-2025-mhumaneval} and MBPP-Bangla~\cite{raihan2025tigercoder}, has begun to address this gap by providing standardized datasets and evaluation protocols. Automated code generation in Bengali is important because it enables more people to access programming tools and resources in their native language, thereby supporting education and local software development.

To advance research, the BLP Workshop\footnote{\url{https://blp-workshop.github.io/}} at IJCNLP-AACL 2025 introduced a shared task on ``Code Generation in Bangla'' \cite{raihan-etal-2025-blp}, providing a benchmark for evaluating models on Bangla instruction-to-Python code generation. This paper contributes to ongoing research, and the key contributions are as follows:

\begin{itemize}
\item Proposed a lightweight and effective system for Bangla-to-Python code generation that combines QLoRA fine-tuning with a feedback-guided inference loop and includes a test-case-aware translation step to ensure semantic alignment with expected input-output behavior.

\item Conducted a systematic evaluation of several open-weight models, including (\texttt{ReasonFlux-Coder-14B}, \texttt{phi-4}, \texttt{Llama-3.1-8B}, \texttt{Qwen3-14B}, \texttt{Qwen2.5-Coder-14B}) on Bangla-to-Python code generation.

\end{itemize}

\section{Related Work}
Recent studies have sought to enhance the robustness of code generation. Python Code Generation by Asking Clarification Questions \cite{li2023python} allows models to query ambiguous prompts, improving correctness through interactive clarification. Self-Debugging \cite{chen2023teaching} proposes a general framework where LLMs iteratively refine their own outputs by leveraging execution feedback, showing improvements across text-to-SQL, code translation, and text-to-Python tasks. Similarly, the Large Language Model Debugger (LDB) \cite{zhong2024debug} incorporates runtime execution signals and block-level debugging, yielding gains on HumanEval \cite{chen2021evaluating} and MBPP \cite{austin2021program}. Building upon recent work in enhancing code generation, efforts to assess LLM capabilities in Bangla have led to the introduction of several benchmarks and resources. For example, the mHumanEval-Bangla dataset \cite{raihan-etal-2025-mhumaneval} extends HumanEval~\cite{chen2021evaluating} to Bangla, while the MBPP-Bangla benchmark~\cite{raihan2025tigercoder} adapts MBPP with crowd-sourced Bangla instructions paired with Python solutions. In parallel, modeling advances include TigerLLM~\cite{raihan-zampieri-2025-tigerllm}, a family of Bangla LLMs outperforming previous open alternatives, and TituLLMs~\cite{nahin-etal-2025-titullms}, which release pretrained Bangla models at 1B and 3B scales with comprehensive benchmarking.

Despite the availability of these benchmarks and improvements, code generation in Bangla remains relatively underexplored. The TigerCoder suite~\cite{raihan2025tigercoder} is an early response to this gap, introducing Bangla-focused multilingual models and benchmarks. This work specifically addresses the task of Bangla-to-Python code generation by proposing a test-driven, feedback-guided refinement approach.

\section{Task and Dataset Descriptions}
The BLP Shared Task-2 tackled the challenge of developing robust Bangla code generation systems. Organizers provided instruction-to-Python code datasets for training (74 samples), development (400 samples), and testing (500 samples). Table \ref{tab:dataset_fields} summarizes the structure and fields of each sample.

\begin{table}[h]
\small
\centering
\begin{tabular}{l p{0.65\columnwidth}}
\hline
\textbf{Field} & \textbf{Description} \\
\hline
id & A unique task identifier. \\
instruction & A Bangla description of the programming task. \\
response & A Python code snippet implementing the task (available only in the training set). \\
test\_list & A list of Python assert statements used for verifying functional correctness during development. \\
\hline
\end{tabular}
\caption{Structure of each dataset sample.}
\label{tab:dataset_fields}
\end{table}

\section{System Description}
\begin{figure}[th]
\centering
\includegraphics[width=.48\textwidth]{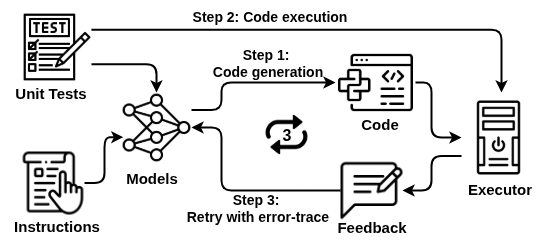}
\caption{Overview of the proposed framework}
\label{fig:feedback-pipeline}
\end{figure}

Figure~\ref{fig:feedback-pipeline} illustrates the overview of our proposed framework.

\subsection{Base Model Benchmarking}
We began by comparing a diverse set of open-weight large language models (LLMs), chosen for their strong performance in code generation and multilingual understanding: 
\texttt{ReasonFlux-Coder-14B} \cite{wang2025co}, \texttt{phi-4} \cite{abdin2024phi}, \texttt{Llama-3.1-8B} \cite{dubey2024llama}, \texttt{codegemma-7b} \cite{team2024codegemma}, \texttt{Qwen3-14B} \cite{yang2025qwen3}, and \texttt{Qwen2.5-Coder-14B} \cite{hui2024qwen2}. 
Each model was tested in a two-shot setting, where we prompted with two Bangla instructions and their reference solutions. 
To evaluate robustness, we repeated the same experiments using translated English instructions. 
Across both setups, \texttt{Qwen2.5-Coder-14B} consistently produced the most reliable and executable Python code, making it the natural choice for our subsequent fine-tuning.

\subsection{Instruction Translation Strategy}
Since the development dataset contained Bangla instructions paired with English test cases, we introduced an LLM-based translation step to better leverage English-centric code models.

For each instruction, the full test suite was included in the prompt, allowing the translator to align the natural language description with the intended input/output behavior. 
This design helped preserve semantic fidelity, especially in tasks where the Bangla phrasing might otherwise be ambiguous. 
We used \texttt{Qwen2.5-Coder-14B} as the translator for the full dataset.

\subsection{Fine-tuning with QLoRA}

Given GPU memory constraints, we fine-tuned the base model using Quantized Low-Rank Adaptation (QLoRA) \cite{dettmers2023qlora}, which combines 4-bit weight quantization with low-rank adapters. 
This approach enabled us to adapt a 14B-parameter model within our compute budget while still leveraging its rich pretraining. 
For fine-tuning, we additionally used the DeepMind MBPP dataset \cite{austin2021program} to supplement our training data. 
We settled on a configuration with rank $r=128$, scaling factor $\alpha=128$, a learning rate of $2 \times 10^{-4}$, and a batch size of 1 with gradient accumulation steps of 4. 
Training was performed for 4 epochs with weight decay of 0.

\subsection{Feedback-guided Inference}
Finally, we introduced a feedback mechanism to improve inference robustness. 
After the model generated a candidate solution, we executed it against all the test cases provided with a 30-second timeout. 
If any test failed, we re-prompted the model with the error trace, allowing it to iteratively refine its output. 
This loop was repeated up to three times, gradually increasing the sampling temperature (0.1 $\rightarrow$ 0.3 $\rightarrow$ 0.5) to encourage diverse candidate solutions. 
We found prompt design to be crucial here, and iteratively refined the error-handling prompt over multiple development runs. 

\section{Experiments}

All experiments were conducted on a single NVIDIA RTX 3090 Ti GPU with 24 GB of memory. 
We used the HuggingFace\footnote{\url{https://huggingface.co/}} \texttt{transformers} library together with the \texttt{PEFT} framework and the \texttt{Unsloth}\footnote{\url{https://unsloth.ai/}} package for efficient training. 
Unless otherwise stated, decoding was performed with a maximum of 768 tokens and an initial temperature of 0.1. 
This setup reflects a modest compute budget, which influenced the design of our training and inference strategies.

\section{Results and Analysis}
System performance was measured using the Pass@1 metric, defined as the proportion of top-1 generated solutions that pass all hidden unit tests. Leaderboard ranking was determined by Pass@1.

\subsection{Few-shot on Bangla Instructions}
We first evaluated all candidate models in a two-shot setting using the original Bangla instructions. 
As shown in Table~\ref{tab:few-shot-bn}, \texttt{Qwen2.5-Coder-14B} and \texttt{Qwen3-14B} emerged as the strongest models, reflecting their multilingual pretraining. 
In contrast, models with weaker multilingual grounding (\texttt{phi-4}, \texttt{codegemma-7b}) struggled to interpret Bangla instructions reliably.

\begin{table}[h]
  \centering
  \begin{tabular}{lc}
    \hline
    \textbf{Model} & \textbf{Pass@1 (Bangla)} \\
    \hline
    ReasonFlux-Coder-14B     & 0.64           \\
    phi-4    & 0.20           \\
    Llama-3.1-8B     & 0.51           \\
    codegemma-7b     & 0.37          \\
    Qwen3-14B      & 0.67           \\
    Qwen2.5-Coder-14B     & \textbf{0.74}            \\\hline
  \end{tabular}
  \caption{Few-shot performance on Bangla instructions.}
  \label{tab:few-shot-bn}
\end{table}

\subsection{Few-shot on Translated Instructions}
Next, we repeated the evaluation after translating Bangla instructions into English using our LLM-based pipeline. 
Table~\ref{tab:few-shot-en} shows that translation substantially boosted performance for most models, with \texttt{Qwen2.5-Coder-14B} again leading at 0.81 Pass@1. 
This demonstrates that while multilingual ability helps, aligning with English-centric pretraining remains advantageous for code generation. Appendix~\ref{sec:bangla-english-tranlation-examples} shows a few examples of Bangla instructions alongside their English translations.

\begin{table}[h]
  \centering
  \begin{tabular}{lc}
    \hline
    \textbf{Model} & \textbf{Pass@1 (English)} \\
    \hline
    ReasonFlux-Coder-14B     & 0.73           \\
    phi-4    & 0.74           \\
    Llama-3.1-8B     & 0.59           \\
    codegemma-7b     & 0.42          \\
    Qwen3-14B      & 0.75           \\
    Qwen2.5-Coder-14B     & \textbf{0.81}            \\\hline
  \end{tabular}
  \caption{Few-shot performance on translated English instructions.}
  \label{tab:few-shot-en}
\end{table}

\subsection{Effect of Fine-tuning}
Fine-tuning \texttt{Qwen2.5-Coder-14B} with QLoRA yielded a notable improvement, raising Pass@1 from 0.81 to 0.90. The best configuration was obtained with rank $r=128$, scaling factor $\alpha=128$, and 0 dropout. We found that any added dropout degraded performance, suggesting that preserving the full training signal was more important than regularization under the limited dataset size. This highlights the value of higher adapter ranks in capturing nuanced mappings from translated instructions to executable logic.

\subsection{Feedback-guided Inference}
Incorporating the feedback mechanism further improved performance to 0.94 Pass@1 (Table~\ref{tab:diff-method}). 
Relative to the few-shot baseline, this represents a +16.05\% absolute improvement, and a +4.44\% gain over fine-tuning alone. 

\begin{table}[h]
  \centering
  \begin{tabular}{lc}
    \hline
    \textbf{Method} & \textbf{Pass@1} \\
    \hline
    QLoRA     & 0.90           \\
    QLoRA + Feedback mechanism    & \textbf{0.94}            \\\hline
  \end{tabular}
  \caption{Impact of feedback-guided inference.}
  \label{tab:diff-method}
\end{table}

Most recoveries came from correcting off-by-one errors, handling edge cases, or aligning outputs with expected formats. However, failures caused by mistranslations or deeper reasoning gaps were rarely resolved, underscoring the limits of inference-time self-correction.
The prompt used for this task is provided in Appendix~\ref{sec:promts}.

\subsection{Translation Error Analysis}
Although the LLM-based translation pipeline preserved semantics in most cases, some instructions suffered from semantic drift. 
A representative example is shown below:

\begin{quote}
\textbf{Original Bangla:} 
{\beng{“একটি চতুর্ভুজ সমীকরণের মূলগুলি একে অপরের প্রতিদ্বন্দ্বিতা কিনা তা পরীক্ষা করার জন্য একটি পাইথন ফাংশন লিখুন”}} \\
\textbf{LLM Translation:} 
“Write a Python function to check whether the roots of two quadratic equations are in competition with each other.” 
\end{quote}

Here, {\beng{“প্রতিদ্বন্দ্বিতা”}} (literally “competition”) was mistranslated as “in competition,” while the intended mathematical meaning was “reciprocal.” 
This caused the generated code to implement incorrect logic. 
Such cases highlight that while test-case-aware prompting improves translation, bridging natural Bangla phrasing with precise programming semantics remains challenging.

\subsection{Shared Task Outcome}
Our final system \texttt{Qwen2.5-Coder-14B} with QLoRA fine-tuning and feedback-guided inference achieved Pass@1 score of 0.934 on the blind evaluation set, ranking second overall in the shared task leaderboard. 
This demonstrates that with careful translation, parameter-efficient fine-tuning, and inference-time self-correction, open-weight LLMs can achieve state-of-the-art performance on Bangla-to-Python code generation.

\section{Conclusion}

This work presents an LLM-based system for generating Bangla-to-Python code. 
The suggested approach integrates an LLM-based translation pipeline, parameter-efficient fine-tuning with QLoRA, and a feedback-guided inference loop. These components enabled the model to achieve Pass@1 accuracies of 0.94 on the development set and 0.934 on the blind test set. These results demonstrate that, with careful translation, efficient adaptation, and test-case-aware inference, LLMs can match the performance of much larger or closed-source systems. The future aim is to utilize fine-tuned, larger-parameter models, such as those employing LoRA and other advanced techniques, to capture more nuanced, task-specific representations. We also aim to evaluate closed-source LLMs available through APIs that benefit from stronger hardware and training pipelines. Improving translation fidelity, particularly for idiomatic Bengali expressions, remains a key challenge and a crucial step toward more robust multilingual code generation.

\section{Limitations}

The system achieved competitive results, but several limitations persist. Methodological challenges such as translation fidelity and error-driven code correction, along with resource constraints in training and deployment, remain. Addressing these issues is essential to enhancing the robustness and scalability of Bangla-to-Python code generation systems. Key limitations include:

\begin{itemize}

\item Hardware constraints restricted us to QLoRA fine-tuning on a single GPU, which prevented exploration of full-precision LoRA or larger models that could offer additional improvements.

\item Although generally effective, the translation pipeline occasionally introduced semantic drift, particularly with idiomatic Bangla expressions and loanwords. These errors affected code generation and were only partially addressed by the feedback mechanism.

\item The feedback-guided inference loop depended on the quality of error traces.
When tracebacks did not identify issues, retries seldom led to meaningful corrections.

\item The experiments were limited to open-weight LLMs, which excluded evaluation of potentially stronger closed-source models that benefit from larger-scale pretraining and inference resources.
\end{itemize}

\bibliography{custom}

@inproceedings{raihan-etal-2025-blp,
     title = "Overview of {BLP}-2025 Task 2: Code Generation in Bangla",
     author = "Raihan, Nishat  and
         Jawad, Mohammad Anas  and
         Rahman, Md Mezbaur  and
         Ulfat, Noshin  and
         Gupta, Pranav  and
         Rahman, Mehrab Mustafy and
         Karmakar, Shubhra Kanti and
         Zampieri, Marcos",
     booktitle = "Proceedings of the Second Workshop on Bangla Language Processing (BLP-2025)",
     month = dec,
     year = "2025",
     publisher = "Association for Computational Linguistics (ACL)"
 }

@inproceedings{raihan-etal-2025-mhumaneval,
     title = "m{H}uman{E}val - A Multilingual Benchmark to Evaluate Large Language Models for Code Generation",
     author = "Raihan, Nishat  and
         Anastasopoulos, Antonios  and
         Zampieri, Marcos",
     editor = "Chiruzzo, Luis  and
         Ritter, Alan  and
         Wang, Lu",
     booktitle = "Proceedings of the 2025 Conference of the Nations of the Americas Chapter of the Association for Computational Linguistics: Human Language Technologies (Volume 1: Long Papers)",
     month = apr,
     year = "2025",
     address = "Albuquerque, New Mexico",
     publisher = "Association for Computational Linguistics",
     url = "https://aclanthology.org/2025.naacl-long.570/",
     doi = "10.18653/v1/2025.naacl-long.570",
     pages = "11432--11461",
     ISBN = "979-8-89176-189-6"
 }

@article{raihan2025tigercoder, title={TigerCoder: A Novel Suite of LLMs for Code Generation in Bangla}, author={Raihan, Nishat and Anastasopoulos, Antonios and Zampieri, Marcos}, journal={arXiv preprint arXiv:2509.09101}, year={2025} }

@article{austin2021program,
  title={Program synthesis with large language models},
  author={Austin, Jacob and Odena, Augustus and Nye, Maxwell and Bosma, Maarten and Michalewski, Henryk and Dohan, David and Jiang, Ellen and Cai, Carrie and Terry, Michael and Le, Quoc and others},
  journal={arXiv preprint arXiv:2108.07732},
  year={2021}
}

@article{wang2025co,
  title={Co-Evolving LLM Coder and Unit Tester via Reinforcement Learning},
  author={Wang, Yinjie and Yang, Ling and Tian, Ye and Shen, Ke and Wang, Mengdi},
  journal={arXiv e-prints},
  pages={arXiv--2506},
  year={2025}
}

@article{abdin2024phi,
  title={Phi-4 technical report},
  author={Abdin, Marah and Aneja, Jyoti and Behl, Harkirat and Bubeck, S{\'e}bastien and Eldan, Ronen and Gunasekar, Suriya and Harrison, Michael and Hewett, Russell J and Javaheripi, Mojan and Kauffmann, Piero and others},
  journal={arXiv preprint arXiv:2412.08905},
  year={2024}
}

@article{dubey2024llama,
  title={The llama 3 herd of models},
  author={Dubey, Abhimanyu and Jauhri, Abhinav and Pandey, Abhinav and Kadian, Abhishek and Al-Dahle, Ahmad and Letman, Aiesha and Mathur, Akhil and Schelten, Alan and Yang, Amy and Fan, Angela and others},
  journal={arXiv e-prints},
  pages={arXiv--2407},
  year={2024}
}

@article{hui2024qwen2,
  title={Qwen2. 5-coder technical report},
  author={Hui, Binyuan and Yang, Jian and Cui, Zeyu and Yang, Jiaxi and Liu, Dayiheng and Zhang, Lei and Liu, Tianyu and Zhang, Jiajun and Yu, Bowen and Lu, Keming and others},
  journal={arXiv preprint arXiv:2409.12186},
  year={2024}
}

@article{yang2025qwen3,
  title={Qwen3 technical report},
  author={Yang, An and Li, Anfeng and Yang, Baosong and Zhang, Beichen and Hui, Binyuan and Zheng, Bo and Yu, Bowen and Gao, Chang and Huang, Chengen and Lv, Chenxu and others},
  journal={arXiv preprint arXiv:2505.09388},
  year={2025}
}

@article{team2024codegemma,
  title={Codegemma: Open code models based on gemma},
  author={Team, CodeGemma and Zhao, Heri and Hui, Jeffrey and Howland, Joshua and Nguyen, Nam and Zuo, Siqi and Hu, Andrea and Choquette-Choo, Christopher A and Shen, Jingyue and Kelley, Joe and others},
  journal={arXiv preprint arXiv:2406.11409},
  year={2024}
}

@inproceedings{li2023python,
  title={Python Code Generation by Asking Clarification Questions},
  author={Li, Haau-Sing Xiaocheng and Mesgar, Mohsen and Martins, Andr{\'e} FT and Gurevych, Iryna},
  booktitle={Proceedings of the 61st Annual Meeting of the Association for Computational Linguistics (Volume 1: Long Papers)},
  pages={14287--14306},
  year={2023}
}

@inproceedings{zhong2024debug,
  title={Debug like a Human: A Large Language Model Debugger via Verifying Runtime Execution Step by Step},
  author={Zhong, Li and Wang, Zilong and Shang, Jingbo},
  booktitle={Findings of the Association for Computational Linguistics ACL 2024},
  pages={851--870},
  year={2024}
}

@article{chen2021evaluating,
  title={Evaluating large language models trained on code},
  author={Chen, Mark and Tworek, Jerry and Jun, Heewoo and Yuan, Qiming and Pinto, Henrique Ponde De Oliveira and Kaplan, Jared and Edwards, Harri and Burda, Yuri and Joseph, Nicholas and Brockman, Greg and others},
  journal={arXiv preprint arXiv:2107.03374},
  year={2021}
}

@inproceedings{raihan-zampieri-2025-tigerllm,
    title = "{T}iger{LLM} - A Family of {B}angla Large Language Models",
    author = "Raihan, Nishat  and
      Zampieri, Marcos",
    editor = "Che, Wanxiang  and
      Nabende, Joyce  and
      Shutova, Ekaterina  and
      Pilehvar, Mohammad Taher",
    booktitle = "Proceedings of the 63rd Annual Meeting of the Association for Computational Linguistics (Volume 2: Short Papers)",
    month = jul,
    year = "2025",
    address = "Vienna, Austria",
    publisher = "Association for Computational Linguistics",
    url = "https://aclanthology.org/2025.acl-short.69/",
    doi = "10.18653/v1/2025.acl-short.69",
    pages = "887--896",
    ISBN = "979-8-89176-252-7"
}

@inproceedings{nahin-etal-2025-titullms,
    title = "{T}itu{LLM}s: A Family of {B}angla {LLM}s with Comprehensive Benchmarking",
    author = "Nahin, Shahriar Kabir  and
      Nandi, Rabindra Nath  and
      Sarker, Sagor  and
      Muhtaseem, Quazi Sarwar  and
      Kowsher, Md  and
      Shill, Apu Chandraw  and
      Ibrahim, Md  and
      Menon, Mehadi Hasan  and
      Muntasir, Tareq Al  and
      Alam, Firoj",
    editor = "Che, Wanxiang  and
      Nabende, Joyce  and
      Shutova, Ekaterina  and
      Pilehvar, Mohammad Taher",
    booktitle = "Findings of the Association for Computational Linguistics: ACL 2025",
    month = jul,
    year = "2025",
    address = "Vienna, Austria",
    publisher = "Association for Computational Linguistics",
    url = "https://aclanthology.org/2025.findings-acl.1279/",
    doi = "10.18653/v1/2025.findings-acl.1279",
    pages = "24922--24940",
    ISBN = "979-8-89176-256-5"
}

@article{chen2023teaching,
  title={Teaching large language models to self-debug},
  author={Chen, Xinyun and Lin, Maxwell and Sch{\"a}rli, Nathanael and Zhou, Denny},
  journal={arXiv preprint arXiv:2304.05128},
  year={2023}
}

@article{dettmers2023qlora,
  title={Qlora: Efficient finetuning of quantized llms},
  author={Dettmers, Tim and Pagnoni, Artidoro and Holtzman, Ari and Zettlemoyer, Luke},
  journal={Advances in neural information processing systems},
  volume={36},
  pages={10088--10115},
  year={2023}
}

\appendix

\section{Bangla-English Instruction Examples}
\label{sec:bangla-english-tranlation-examples}

Table~\ref{tab:translation-examples} illustrates representative Bangla instructions 
from the dataset alongside their English translations used for code generation.

\begin{table}[h]
\centering
\renewcommand{\arraystretch}{1.2}
\begin{tabular}{p{0.45\linewidth} p{0.45\linewidth}}
\hline
\textbf{Bangla Instruction} & \textbf{English Translation} \\
\hline
{\beng{প্রদত্ত পর্যায়ক্রমিক ফাংশনের জন্য সর্বনিম্ন সম্ভাব্য মান খুঁজে পেতে একটি পাইথন ফাংশন লিখুন।}} & 
Write a Python function to find the minimum possible value for a given arithmetic sequence. \\
\hline
{\beng{প্রদত্ত তালিকায় টুপল বৈশিষ্ট্য হিসাবে রেকর্ড তালিকার সর্বাধিক মান খুঁজে পেতে একটি ফাংশন লিখুন।}} & 
Write a function to find the maximum value in each tuple's list within a given list and return the tuples with their respective maximum values. \\
\hline
{\beng{মানচিত্র এবং ল্যাম্বদা ফাংশন ব্যবহার করে দুটি তালিকা ভাগ করার জন্য একটি ফাংশন লিখুন।}} & 
Write a function to divide two lists element-wise using map and lambda functions. \\
\hline
{\beng{প্রদত্ত স্ট্রিং-এর সমস্ত স্পেসকে অক্ষর * তালিকা আইটেম * তালিকা আইটেম * তালিকা আইটেম * তালিকা আইটেম '\%20' দিয়ে প্রতিস্থাপনের জন্য একটি ফাংশন লিখুন।}} & 
Write a function to replace all spaces in a given string with '\%20'. \\
\hline
{\beng{একটি মিশ্র তালিকা থেকে এমনকি সংখ্যা খুঁজে পেতে একটি পাইথন ফাংশন লিখুন।}} & 
Write a Python function to extract numbers from a mixed list. \\
\hline
\end{tabular}
\caption{Examples of Bangla instructions and their English translations.}
\label{tab:translation-examples}
\end{table}

\section{Prompts}
\label{sec:promts}

This appendix contains the prompts employed in system development and testing, offering additional insight into our framework.
\subsection{Translation Prompt}

\begin{lstlisting}
messages = [
    {
        "role": "system",
        "content": """You are a professional translator specializing in technical and programming content...
        
Translation Guidelines:
- Preserve Technical Accuracy: Maintain exact meaning of programming concepts
- Keep Code Elements Intact: Preserve English technical terms in standard form
- Maintain Instructional Clarity: Ensure natural English coding instructions
- Precision Over Literal Translation: Focus on exact intended meaning

Example:
Input: (*@{\beng{একটি স্ট্রিং-এ একটি আক্ষরিক স্ট্রিং অনুসন্ধান করার জন্য একটি ফাংশন লিখুন এবং রেজেক্স ব্যবহার করে মূল স্ট্রিং-এর মধ্যে অবস্থানটি খুঁজে বের করুন যেখানে প্যাটার্নটি ঘটে।}}@*)
Output: Write a function to search for a literal string within a main string and find the position within the main string where the pattern occurs using regex."""
    },
    {
        "role": "user",
        "content": f"Translate this Bangla coding instruction to English: {bangla_instruction}"
    }
]
\end{lstlisting}

\subsection{Few-shot Prompt (Bangla Instructions)}

\captionsetup{type=lstlisting}
\captionof{lstlisting}{Few-shot prompt with translated English examples}
\begin{lstlisting}[language=Python, basicstyle=\ttfamily\small, breaklines=true]
system_message = """You are an expert Python programmer. Your task is to generate clean, efficient, and correct Python functions that pass all given test cases.

CRITICAL RULES:
1. ALWAYS wrap your code in ```python ``` blocks
2. Write ONLY the function implementation, no extra explanations
3. Use the EXACT function name from the example
4. Ensure the function passes ALL test cases
5. Handle edge cases and invalid inputs appropriately
6. Use appropriate data types based on test case patterns

Here are examples of how to solve different types of problems:

EXAMPLE 1 - String Processing:
Task: (*@{\beng{একটি প্রদত্ত স্ট্রিং-এ প্রথম পুনরাবৃত্ত অক্ষর খুঁজে পেতে একটি পাইথন ফাংশন লিখুন।}}@*)
Test Cases:
assert first_repeated_char("abcabc") == "a"
assert first_repeated_char("abc") == "None"
assert first_repeated_char("123123") == "1"

Expected Solution:
```python
def first_repeated_char(s):
    seen = set()
    for char in s:
        if char in seen:
            return char
        seen.add(char)
    return "None"
```
EXAMPLE 2 - Mathematical Function:
Task: (*@{\beng{প্রদত্ত পূর্ণসংখ্যাটি একটি মৌলিক সংখ্যা কিনা তা পরীক্ষা করার জন্য একটি ফাংশন লিখুন।}}@*)
Test Cases:
assert prime_num(13) == True
assert prime_num(7) == True
assert prime_num(-1010) == False

Expected Solution:
```python
def prime_num(n):
    if n < 2:
        return False
    if n == 2:
        return True
    if n % 2 == 0:
        return False
    for i in range(3, int(n**0.5) + 1, 2):
        if n % i == 0:
            return False
    return True
```
Code Quality Standards:

- Write code with proper indentation
- Optimize for correctness first, then efficiency
- Handle common edge cases (empty inputs, None values, negative numbers, etc.)

Return the exact data type shown in test cases"""

user_prompt = f"""Generate a Python function for this problem:

Task: {instruction}

Test Cases:
{test_list}

Expected Function Name: {function_name}

Requirements:

- Follow the examples shown in the system message
- Analyze the test cases carefully to understand input/output patterns
- Implement the function to pass ALL test cases exactly
- Return the appropriate data type as shown in test cases
- Handle edge cases gracefully (empty inputs, invalid values, etc.)
- Use efficient algorithms where applicable

Generate ONLY the Python function wrapped in python blocks. No explanations needed."""
\end{lstlisting}

\subsection{Few-shot Prompt (Translated Instructions)}
\captionsetup{type=lstlisting}
\captionof{lstlisting}{Few-shot prompt with translated English examples}
\begin{lstlisting}[language=Python, basicstyle=\ttfamily\small, breaklines=true]
system_message = """You are an expert Python programmer. Your task is to generate clean, efficient, and correct Python functions that pass all given test cases.

CRITICAL RULES:
1. ALWAYS wrap your code in ```python ``` blocks
2. Write ONLY the function implementation, no extra explanations
3. Use the EXACT function name from the example
4. Ensure the function passes ALL test cases
5. Handle edge cases and invalid inputs appropriately
6. Use appropriate data types based on test case patterns

Here are examples of how to solve different types of problems:

EXAMPLE 1 - String Processing:
Task: Write a Python function to find the first repeated character in a given string.
Test Cases:
assert first_repeated_char("abcabc") == "a"
assert first_repeated_char("abc") == "None"
assert first_repeated_char("123123") == "1"

Expected Solution:
```python
def first_repeated_char(s):
    seen = set()
    for char in s:
        if char in seen:
            return char
        seen.add(char)
    return "None"
```
EXAMPLE 2 - Mathematical Function:
Task: Write a function to check if a given integer is a prime number.
Test Cases:
assert prime_num(13) == True
assert prime_num(7) == True
assert prime_num(-1010) == False

Expected Solution:
```python
def prime_num(n):
    if n < 2:
        return False
    if n == 2:
        return True
    if n % 2 == 0:
        return False
    for i in range(3, int(n**0.5) + 1, 2):
        if n % i == 0:
            return False
    return True
```
Code Quality Standards:

- Write code with proper indentation
- Optimize for correctness first, then efficiency
- Handle common edge cases (empty inputs, None values, negative numbers, etc.)

Return the exact data type shown in test cases"""

user_prompt = f"""Generate a Python function for this problem:

Task: {instruction}

Test Cases:
{test_list}

Expected Function Name: {function_name}

Requirements:

- Follow the examples shown in the system message
- Analyze the test cases carefully to understand input/output patterns
- Implement the function to pass ALL test cases exactly
- Return the appropriate data type as shown in test cases
- Handle edge cases gracefully (empty inputs, invalid values, etc.)
- Use efficient algorithms where applicable

Generate ONLY the Python function wrapped in python blocks. No explanations needed."""
\end{lstlisting}

\subsection{Feedback/Retry Prompt}
\captionsetup{type=lstlisting}
\captionof{lstlisting}{Few-shot prompt with translated English examples}
\begin{lstlisting}[language=Python, basicstyle=\ttfamily\small, breaklines=true]
attempt_analysis = f"""

# PATTERN ANALYSIS FROM {len(previous_attempts)} ATTEMPTS:
- Attempt 1: {len(previous_attempts[0])} characters
- Latest: {len(previous_attempts[-1])} characters
- Different approaches tried: {len(set(attempt[:50] for attempt in previous_attempts))}

# AVOID REPEATING: The same logic pattern has failed multiple times. Try a fundamentally different approach."""

    # Enhanced error analysis based on error type
    specific_guidance = ""
    if failed_test['status'] == 'ASSERTION_FAILED':
        specific_guidance = """
## ASSERTION FAILURE GUIDANCE:
- Check return data type (int, str, list, tuple, bool)
- Verify exact return format matches expected output
- Consider sorting if order doesn't matter
- Handle empty cases explicitly"""
    elif failed_test['status'] == 'RUNTIME_ERROR':
        error_msg = failed_test['error'].lower()
        if 'index' in error_msg or 'list' in error_msg:
            specific_guidance = """
## INDEX/LIST ERROR GUIDANCE:
- Check for empty list/string handling
- Verify array bounds (0 to len-1)
- Handle edge case when input is empty"""
        elif 'key' in error_msg or 'dict' in error_msg:
            specific_guidance = """
## DICTIONARY ERROR GUIDANCE:
- Check if key exists before accessing
- Use .get() method with default values
- Initialize dictionaries properly"""
        elif 'attribute' in error_msg:
            specific_guidance = """
## ATTRIBUTE ERROR GUIDANCE:
- Check object types before method calls
- Verify variable is initialized
- Import necessary modules"""
    
    # Create comprehensive error feedback
    feedback = f"""

## PREVIOUS ATTEMPT FAILED - ADVANCED DEBUGGING:

- Error Type: {failed_test['status']}
- Error Message: {failed_test['error']}
- Failing Test Case: {failed_test['test_case']}
- Failed at Test #{failed_test['index']} out of {len([r for r in test_results if 'index' in r])}

{specific_guidance}

{attempt_analysis}

# SYSTEMATIC DEBUGGING APPROACH:
1. ANALYZE INPUT/OUTPUT: What data types and patterns do test cases show?
2. EDGE CASE CHECK: Empty inputs, single elements, boundary values
3. ALGORITHM CHOICE: Is this DP, greedy, two-pointer, sliding window, etc.?
4. IMPLEMENTATION: Step through the failing test case manually
5. IMPORTS: Add math, re, collections, itertools if needed

# Original Task: {instruction}

# CRITICAL SUCCESS FACTORS:
- Function signature must match test case exactly
- Return type must match expected output precisely  
- Handle ALL edge cases shown in test patterns
- Use efficient algorithm for the problem type

GENERATE A COMPLETELY NEW APPROACH - Previous attempts failed for a reason."""
\end{lstlisting}

\end{document}